\pdfoutput=1

\documentclass[11pt]{article}

\usepackage[]{EMNLP2023}

\usepackage{times}
\usepackage{latexsym}

\usepackage[T1]{fontenc}

\usepackage[utf8]{inputenc}

\usepackage{microtype}

\usepackage{amsfonts} 
\usepackage{booktabs} 
\usepackage{multirow} 
\usepackage{subfigure} 
\usepackage{xspace}
\usepackage{graphicx}
\usepackage{soul}
\usepackage{color,xcolor}
\usepackage{colortbl}
\usepackage[english]{babel}
\usepackage{balance}
\usepackage{amsmath,bm}
\usepackage{algorithm}
\usepackage{algpseudocode}
\usepackage{makecell}
\usepackage{float}
\usepackage{CJKutf8}
\usepackage{hyperref}
\definecolor{Gray}{gray}{0.9}
\usepackage{inconsolata}

\title{Only One Relation Possible?\\Modeling the Ambiguity in Event Temporal Relation Extraction}

\author{
    Yutong Hu$^{1,2}$, 
    Quzhe Huang$^{1,2}$, 
    \textbf{Yansong Feng}$^{1}$\thanks{\;\;Corresponding author.}~~ \\
    $^1$Wangxuan Institute of Computer Technology, Peking University, China \\ 
    $^2$School of Intelligence Science and Technology, Peking University \\
    {\tt \{huangquzhe,huyutong,fengyansong,\}} 
     {\tt @pku.edu.cn} \\
}

\begin{document}
\maketitle
\begin{abstract}
Event Temporal Relation Extraction (ETRE) aims to identify the temporal relationship between two events, which plays an important role in natural language understanding. Most previous works follow a single-label classification style,
classifying an event pair into either a specific temporal relation (e.g., \textit{Before}, \textit{After}), or a special label \textit{Vague} when there may be multiple possible temporal relations between the pair. 
In our work, instead of directly making predictions on \textit{Vague}, we propose a multi-label classification solution for ETRE (METRE) to infer the possibility of each temporal relation independently, where we treat \textit{Vague} as the cases when there is more than one possible relation between two events. 
We design a speculation mechanism to explore the possible relations hidden behind \textit{Vague}, which enables the latent information to be used efficiently. 
Experiments on TB-Dense, MATRES and UDS-T show that our method can effectively utilize the \textit{Vague} instances to improve the recognition for specific temporal relations and outperforms most state-of-the-art methods. 
\end{abstract}

\section{Introduction}
\label{sec:intro}
Event Temporal Relation Extraction (ETRE) is a task to determine the temporal relationship between two events in a given text, which could benefit many downstream tasks, such as summarization, generation, question answering and so on \cite{ng-etal-2014-exploiting, yu-etal-2017-improved, shi-etal-2019-unsupervised}.

According to the ambiguity of the temporal relationship between an event pair, most benchmarks categorize the temporal relationship between events into two types: the well-defined relations \cite{cassidy2014annotation} and \textit{Vague}. Well-defined relations indicate the specific temporal relations that can be explicitly positioned in the timeline. For example, TB-Dense \cite{cassidy2014annotation} contains five well-defined relations: \textit{Before}, \textit{After}, \textit{Include}, \textit{Is Included}, \textit{Simultaneous}, while MATRES \cite{ning2018multi} have three: \textit{Before}, \textit{After} and \textit{Equal}. As for the special label, \textit{Vague}, it indicates the cases when people, based on the currently available context, can not make an agreement on which well-defined relation should be chosen for the given event pair. Due to its complexity and ambiguity, how to 
better characterize \textit{Vague} and other well-defined relations is still a challenge for ETRE. 


\begin{table}[t!]
    \centering
    \small
    \begin{tabular}{l|c}
        \toprule
        \multicolumn{2}{l}{\makecell[l]{\textbf{TEXT:} \\My son has fallen asleep $(e_1)$, so I have some free time\\to read $(e_2)$ for a while.}}\\
        \midrule
        \textbf{Possible Scenario 1:} & \textbf{Possible Relation 1} \\
        \makecell[l]{My son woke up while I was\\reading.}  & \textit{Before}\\
        \midrule
        \textbf{Possible Scenario 2:} & \textbf{Possible Relation 2} \\
        \makecell[l]{My son didn't wake up until I\\finished reading the book.}  & \textit{Include}\\
        \bottomrule
    \end{tabular}
    \caption{An example of temporal relation \textit{Vague} between $(e_1, e_2)$. Event "asleep" could happen either before or include event "read".}
    \label{tab:example}
    \vspace{-0.3cm}
\end{table}

Although recent works have made many attempts to reduce the inconsistency of temporal relation labeling, such as using course relations \cite{verhagen2007semeval} or setting rules to 
push annotators to carefully consider \textit{Vague}~\cite{ning2018multi}, 
they still pay less attention to why various \textit{Vague} cases are challenging and how to deal with them better.
%
For example, in Table~\ref{tab:example}, given the text "My son has fallen asleep, so I have some free time to read for a while.", we can only infer that event "asleep" happening before or include event "read" are both possible, and then determine event pair <asleep, read> as \textit{Vague}. In other words,  \textit{Vague} can be explained as an event pair with more than one possible temporal relationship between them. 

Most previous studies on ETRE ignore the premise that \textit{Vague} is determined due to the multiple possible well-defined temporal relations between an event pair, and treat \textit{Vague} independently and equally as other well-defined relations. They formulate ETRE as a single-label classification task, in which they directly predict by choosing the most possible one from all candidate relations.
However, such a single-label classification paradigm will cause the model's confusion between the meaning of \textit{Vague} and other potentially related well-defined relations. Take the sentence in Table~\ref{tab:example} as an example again; the model may capture the information that there is the possibility of event "asleep" happening before event "read", thus wrongly classifying <asleep, read> into \textit{Before}. On the other hand, after being told that this case should be considered as \textit{Vague}, the model will punish all features contributing to \textit{Before}. Unfortunately, doing this will actually confuse the model, since the sentence contains clues for both \textit{Before} and \textit{Include}, the reason why annotators label it with \textit{Vague}.  

To address this problem, instead of treating \textit{Vague} as a single label independent from others, we view \textit{Vague} as the situation when there is more than one possible well-defined temporal relation between an event pair. We propose a Multi-label Event Temporal Relation Extraction (METRE) method that transforms the task into a multi-label classification style, and makes predictions on every well-defined relation about its possibility being the temporal relation between the given event pair. If more than one relation is of high possibility, we consider the relation between the event pair as \textit{Vague}. Thus the intrinsic composition of \textit{Vague} can be properly formulated and make it distinguishable from other well-defined relations.

Experiments on TBDense, MATRES and UDS-T \cite{vashishtha-etal-2019-fine} show METRE could outperform most
previous state-of-the-art methods, indicating our model could better characterize and make full use of the special label \textit{Vague}. Further analysis 
demonstrates that, instead of the tendency to predict an event pair as \textit{Vague}, our model predicts more well-defined relations with higher accuracy 
compared to the baseline model. 
Consistent improvement in low-data scenarios also shows that we can explore the information hidden behind \textit{Vague} 
efficiently. We also show that our method could even find those well-defined relations that possibly cause the \textit{Vague} annotations, providing further interpretability for the ETRE task.

\section{Related Work}
Disagreement is widespread during the temporal relation annotation process. Many previous works, such as TempEval-3 \cite{uzzaman2013semeval} and THYME \cite{styler-iv-etal-2014-temporal}, annotate every event pair by at least two annotators, and the final result is obtained with the adjudication of the third annotator. Similarly, MAVEN-ERE \cite{wang2022maven} uses majority voting to produce the final gold standard annotations. These datasets only contain well-defined relations after achieving agreement among annotators. Many studies retain controversial information in the corpus. For example, TB-Dense \cite{cassidy2014annotation} treat the relation between an event pair as \textit{Vague} if more than half of the annotators disagree on certain well-defined relations, while \citet{verhagen2007semeval} and MATRES \cite{ning2018multi} directly provide relation \textit{Vague} for each annotator to choose. Differently, UDS-T \cite{vashishtha-etal-2019-fine} preserves the detailed annotation results of every annotator and flexible rules of the disagreement adjudication 
can be applied to it. 

The ambiguity of \textit{Vague} is often neglected in ETRE. Previous studies on temporal relation extraction are mainly formulated as a single-label classification task, overlooking the intrinsic nature of \textit{Vague} and treating every relation equally. They pay attention to the essential context information extraction and are looking forward to obtaining a better event pair representation. 
With the rapid development of pre-trained language models, most recent works \cite{han2019joint,wang2020joint} adopt BERT \cite{devlin-etal-2019-bert} or RoBERTa \cite{liu2019roberta} to obtain contextualized representations for event pairs. Besides, several researchers \cite{meng-etal-2017-temporal,cheng-miyao-2017-classifying,DBLP:journals/corr/abs-2104-09570} notice the importance of syntactic features, e.g., Part-Of-Speech tags, dependency parses, etc., and incorporate them into the model. Recent studies try to model temporal relations by taking their definitions and properties into consideration.  
\citet{wen-ji-2021-utilizing} predict the relative timestamp in the timeline, and \citet{hwang-etal-2022-event} express the temporal information through box embeddings with the use of the symmetry and conjunctive properties. Nevertheless, the complex information in \textit{Vague} can not be well represented only as a single label. \citet{wang2022extracting} regard \textit{Vague} as a source of distributional uncertainty and incorporate Dirichlet Prior \cite{malinin2018predictive, malinin2019reverse} to provide the predictive uncertainty estimation. However, they ignore the semantic information relevance between \textit{Vague} and other well-defined relations, as the example shown in Table~\ref{tab:example}. Different from previous studies, we adopt a simple event pair representation encoder, and shift our focus to the classification module to better reflect the unique nature of different relations.
After scrutinizing the definition of \textit{Vague}, we employ a multi-label classification-based approach to capture its intrinsic ambiguity among several possible well-defined relations, thereby enhancing our model's comprehension of all temporal relations.










\section{Our Approach}
\begin{figure*}[pt]
    \center
    \includegraphics[width=0.9\textwidth]{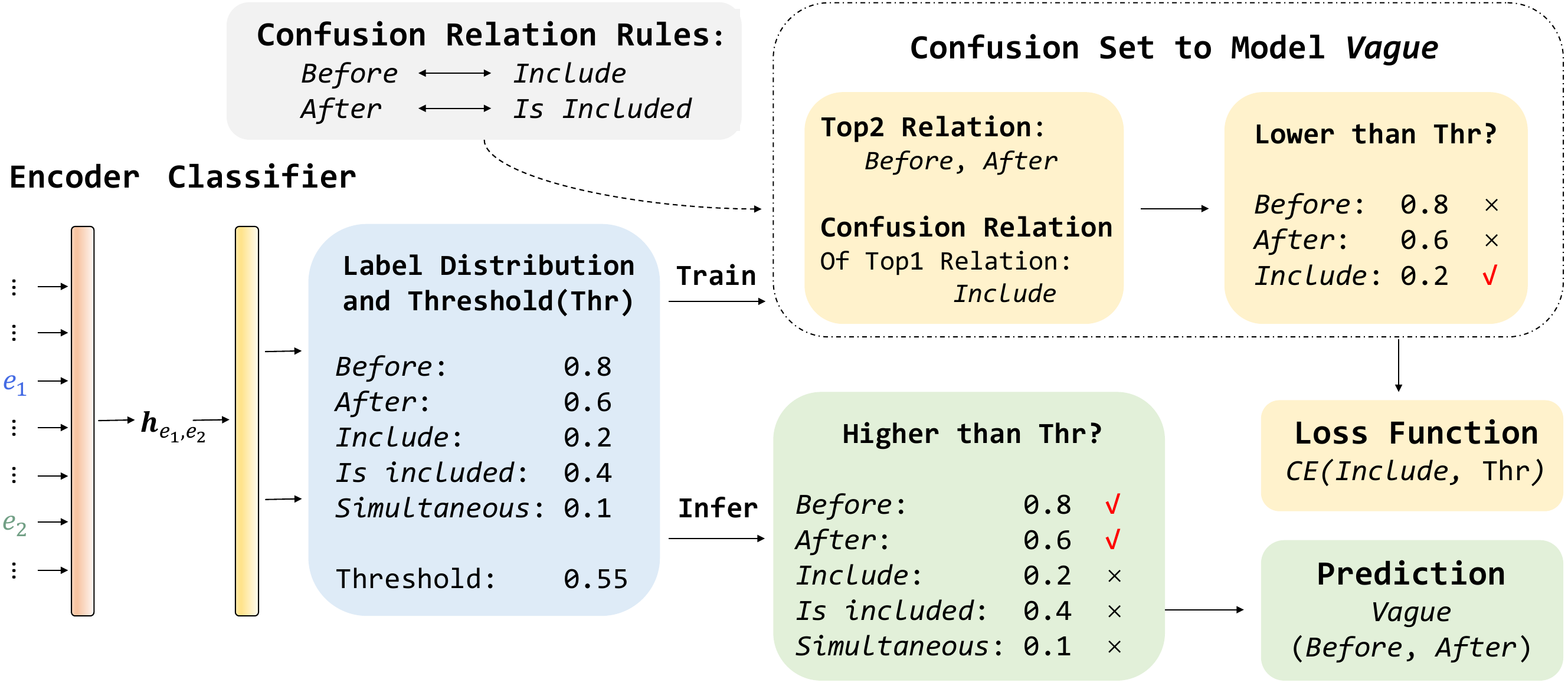}
    \caption{The overview of METRE. $h_{e_1,e_2}$ is the representation of the event pair $(e_1,e_2)$.}
    \label{fig:architecture}
\end{figure*}


 Given an input text sequence $\mathbf{S} = [w_1, w_2, ..., w_n]$ and an event pair $(e_1, e_2)$, where both $e_1$ and $e_2$ are in $\{w_i|i\in[1, n]\}$. The task of temporal relation extraction is to predict the temporal relation $r\in \tilde{R} = R\cup \{Vague\}$ between $e_1$ and $e_2$, where $R$ is the set of well-defined temporal relations.  
 As shown in Figure~\ref{fig:architecture}, METRE includes an encoder to get event representations, a classifier to obtain the probability distribution $P(R)$ and the threshold value $T$. We define Confusion Set (CS) to simulate the label composition of \textit{Vague} and use it to train our model. Finally, we obtain the prediction with our multi-label classifier. If more than one relation is chosen, our model will output \textit{Vague}. 

\subsection{Encoder Module}
In our work, we adopt pre-trained language models BERT \cite{devlin-etal-2019-bert} and RoBERTa \cite{liu2019roberta} as the encoder module. Taking the text sequence $\mathbf{S}$ and an event pair $(e_1, e_2)$ as input, it computes the contextualized representation for the event pair. Following \citet{zhong2021frustratingly}, we insert typed markers to highlight two event mentions in the given text at the input layer, i.e.,  <E1:L> $e_1$ <E1:R>, and <E2:L> $e_2$ <E2:R>. 
Additionally, inspired by \citet{DBLP:conf/emnlp/WaddenWLH19}, we add cross-sentence context, i.e., one sentence before and after the given text, expecting pre-trained language models to capture more comprehensive information. 
We feed the new sequence $\tilde{\mathbf{S}}$ into the encoder and obtain the event pair representation:
\begin{equation}
    \mathbf{h}_{e_1,e_2} = MLP_1([\mathbf{x}_{\hat{e_1}}||\mathbf{x}_{\hat{e_2}}])
    \label{eq:event_pair}
\end{equation}
where $\mathbf{x}$ is the output representation, $\hat{e_i}$ is the indices of $e_i$ in $\tilde{\mathbf{S}}$, [·||·] is the concatenation operator, and $MLP_1$ is a multilayer perceptron.

\subsection{Multi-Label Classifier}



Different from single-label classifications, which directly choose the relation of the highest probability from $\tilde{R}$, we adopt a multi-label classifier to better characterize \textit{Vague}. Taking the event pair representation $\mathbf{h_{e_1,e_2}}$ from encoder as input, our classifier will calculate the probability distribution over well-defined relation set $R$ and a threshold value $T$. Inspired by \citet{zhou2020documentlevel}, we adopt a learnable, adaptive threshold for our classifier to make decisions. Therefore, we obtain the probability distribution $P(R|e_1,e_2)$ and $T_{e_1,e_2}$ with:
\begin{equation}
    P(R|e_1, e_2) = MLP_2(\mathbf{h}_{e_1,e_2})
\end{equation}
\begin{equation}
    T_{e_1,e_2} = MLP_3(\mathbf{h}_{e_1,e_2})
\end{equation}
where $P(R|e_1, e_2)\in \mathbb{R}^{|R|}$ and $T_{e_1,e_2}\in \mathbb{R}$.

Therefore, the temporal relationship of event pair $(e_1,e_2)$ can be deduced from $P(R|e_1,e_2)$ and $T_{e_1,e_2}$. Specifically, if only one relation whose probability is higher than the threshold, we consider this well-defined relation as the final prediction. While if more than one relation's probability is higher than the threshold, we think the temporal relation of the event pair is \textit{Vague}, and we can provide detailed information on exactly what relations cause the ambiguity. As the example shown in Figure~\ref{fig:architecture}, we infer the temporal relation as \textit{Vague} due to the high possibility of both relation \textit{Before} and \textit{After}. There are also some cases that all relations' probability are lower than the threshold,  and we map it to \textit{Vague} as well, which can be explained as the model does not have enough confidence in any relation for the event pair.



\subsection{Training Process}

When training the multi-label classifier, we need the gold labels of each event pair to calculate the training loss. However, this is not trivial in our case. For the instances labeled as \textit{Vague}, we do not exactly know their original possible well-defined relation labels, e.g., \textit{Before} and \textit{Include} for the example sentence in Table~\ref{tab:example}. Therefore, we need a new mechanism to identify those possible well-defined relations behind  \textit{Vague} to make sure proper training for our model, i.e., providing a proper reward to those original well-defined relation labels which result in \textit{Vague}. In our approach, we first utilize $P(R)$ to speculate a dynamic Confusion Set, which consists of the potential relations underlying \textit{Vague}. Then relations in Confusion Set are regarded as gold labels of \textit{Vague} at the current stage, and are used in training loss calculation.

\subsubsection{Construction of Confusion Set}
\label{sec:pos_1}
To address the aforementioned problem, we construct a \textbf{Confusion Set} ($CS$) by speculating some of the most likely well-defined relations between the event pair. 
Considering what we have during training, we think there could be the following two sources to build a dynamic CS to formulate \textit{Vague}:

\paragraph{Top2 Relations.} 
After a period of training in well-defined relations, we can consider that our model has the basic ability to provide a relatively accurate probability distribution over well-defined temporal relations for an event pair. Since \textit{Vague} indicates that there are at least two temporal relations that may exist between an event pair, we, therefore, adopt the top two well-defined relations, $r_{fir}$ and $r_{sec}$, ranked by the current classifier as the possible relations between the event pair. 

\begin{table}[t!]
    \centering
    \scriptsize
    \begin{tabular}{c|ccccc}
        \toprule
        r & \textit{Before} & \textit{After} & \textit{Include} & \textit{Is included} & \textit{Simultaneous} \\
        \midrule
        \makecell[l]{$\bar{r}$} & \textit{Include} & \textit{Is included} & \textit{Before} & \textit{After} & - \\
        \bottomrule
    \end{tabular}
    \caption{Confusion Relation Mapping Rules.}
    \label{tab:confusion relation}
    \vspace{-0.3cm}
\end{table}

\paragraph{Top1's Confusion relation.}
The determination of a temporal relation is always based on the relations between the start-point and end-point. \citet{ning2018multi} claim that comparisons of end-points, i.e., $t_{end}^1$ vs. $t_{end}^2$, are more difficult than comparing start-points i.e., $t_{start}^1$ vs. $t_{start}^2$), which can be attributed to the ambiguity of expressing and perceiving duration of events \cite{COLLFLORIT201141}. Accordingly, we think that \textit{Vague} may come from the ambiguity between the end-points relation of an event pair. For example, if it is easy to figure out $e_1$ starting before $e_2$ while hard to decide the end-point relation, annotators will have disagreement between \textit{Before} ($t_{start}^1<t_{start}^2\land t_{end}^1<t_{end}^2$) and \textit{Include} ($t_{start}^1<t_{start}^2\land t_{end}^1\geq t_{end}^2$). Here we call \textit{Before} and \textit{Include} as \textbf{confusion relations} to each other. According to the ambiguity of end-points, Table~\ref{tab:confusion relation} shows all relations $r\in R$ together with their confusion relations $\bar{r}$. 

Therefore, back to our case, 
when a warmed-up classifier outputs a probability distribution over well-defined relations for a \textit{Vague} instance, we guess that the top-ranked relation $r_{fir}$ along with its confusion counterpart $\bar{r}_{fir}$ are most likely to cause the \textit{Vague} label. 

In summary, we think \textit{Vague} most likely consists of these three relations: $CS=\{r_{fir}, \bar{r}_{fir}, r_{sec}\}$. Notice that, $CS$ comes from the probability distribution provided by a warmed-up classifier, and can not fully represent the intrinsic composition of \textit{Vague}. To avoid error accumulation, if the probability of $r_i$ is higher than the threshold $T$ during training, we will not give it a further reward. Hence, we define $CS_{T}=\{r_i|r_i\in CS, P(r_i)<T\}$.

\subsubsection{Training Objective}
For the instance with well-defined relation $r\in R$ as the gold label, all other well-defined relations are impossible to be the temporal relation between the given event pair. Therefore, we reward the gold label $r$ and penalize the probability of other relations, respectively. The loss functions are formulated as:
\begin{equation}
    \small
    L_1 = -log(\frac{exp(P(r))}{exp(P(r))+exp(T)}) \\
\end{equation}
\begin{equation}
\label{eq:loss2}
    \small
    L_2 = -log(\frac{exp(T)}{\sum_{r_i\in R\backslash \{r\}}exp(P(r_i))+exp(T)})
\end{equation}

While for \textit{Vague}, considering that some potential relations may be ignored and not to be considered into $CS$, we avoid penalizing the probability of any other relations. Besides, as a warmed-up classifier, the model provides nearly random probability distribution at the first few steps of the training process, so we set a linear increasing weight $w$ to control the effect of the loss from \textit{Vague}. Thus we have:
\begin{equation}
    \small
    L_3 = -w*\sum_{r_i\in CS_T}log(\frac{exp(P(r_i))}{\sum_{r_i^{'}\in CS_T}exp(P(r_i^{'}))+exp(T)}) \\
\end{equation}
where $w = MIN(\alpha*t*\bar{w}, \bar{w})$. $\alpha$ is the increasing rate and $t$ is the training step. $w$ will not exceed $\bar{w}$ due to the uncertainty in $CS$.
Both $\bar{w}$ and $\alpha$ are hyperparameters. And the final loss function is:
\begin{equation}
    \small
    L = L_1 + L_2+L_3
\end{equation}

\section{Experimental Setup}

\subsection{Dataset}

We evaluate our model in three public temporal relation extraction datasets. Detailed data statistics of each dataset are reported in Appendix~\ref{app:data statistics}.

\paragraph{TB-Dense}
TB-Dense takes both start-point and end-point relations into consideration, and defines 5 well-defined temporal relations and \textit{Vague}, which accounts for 49.9\% in its training set. We split the dataset into train/dev/test sets following the previous works~\cite{han2019joint,wen-ji-2021-utilizing}.
\paragraph{MATRES}
MATRES focuses on the start-points relation and reduces the temporal relations into 3 well-defined relations. The proportion of \textit{Vague} is decreased to 12.2\% in MATRES. We use the same split of train/dev/test sets as \citet{han2019joint}.

\paragraph{UDS-T}
\label{sec:udst}
Instead of explicitly annotating the temporal relation
UDS-T \cite{vashishtha-etal-2019-fine}, asks annotators to determine the relative timeline between event pairs. Following the definition in TB-Dense, we obtain the temporal relation of each event pair in UDS-T. Every event pair in the validation and test set is labeled by 3 annotators, and temporal relation is determined by a majority voting. \textit{Vague} is labeled if 3 annotators all disagree with each other. With \textit{Vague} in the validation and test set, we merge them and designate 80\% of it as a new training set, 10\% as new validation set and test set respectively, where \textit{Vague} accounts for 23.9\% in the new training set. The details of UDS-T are shown in Appendix~\ref{app:udst map}.

\subsection{Data Enhancement}
Following \citet{DBLP:journals/corr/abs-2104-09570}, we utilize the symmetry property of temporal relations to expand our training set for data enhancement. For example, if the temporal relation of an event pair $(e_1, e_2)$ is \textit{Before}, we add the event pair $(e_2, e_1)$ with relation \textit{After} into the training set. We do not expand the validation set and test set for a fair comparison. 

\subsection{Evaluation Metrics}
We use micro-F1 score as the evaluation metric following the previous works \cite{DBLP:journals/corr/abs-2104-09570,mathur-etal-2021-timers}. 
Besides, we implement a baseline model, which adopts the same encoder as our model, and use an MLP as a single-label classifier over $\tilde{R}$, for a better comparison of our approach.

\section{Experiment}

\begin{table*}[t]
    \centering
    \small
    \begin{tabular}{c|l|c|c|c}
        \toprule[2pt]
        \bf PLM & \makecell[c]{\bf Model} & \bf TB-Dense & \bf MATRES & \bf UDST \\
        \midrule
        \multirow{5}{*}{\makecell[c]{Base}} & HNP \citet{han2019joint}&  64.5 &  75.5 & - \\
        & BERE \cite{hwang-etal-2022-event} & - & 77.3 & - \\
        & Syntactic \cite{DBLP:journals/corr/abs-2104-09570}* & 66.7 & 79.3 & - \\
        \cmidrule{2-5}
        & Enhanced-Baseline & 65.3 & 77.7 & 51.3 \\
        & METRE(Ours)  & 67.9  & 79.2  & 52.3 \\
        \midrule[1.4pt]
        \multirow{7}{*}{\makecell[c]{Large}} & Syntactic \cite{DBLP:journals/corr/abs-2104-09570}*  & 67.1 & 80.3 & - \\
        & Time-Enhanced \cite{wen-ji-2021-utilizing} & - & 81.7 & - \\
        & SCS-EERE \cite{Man_Ngo_Van_Nguyen_2022}\dag  & - & 81.6 & - \\
        & TIMERS \cite{mathur-etal-2021-timers} & 67.8 & 82.3 & - \\
        & Faithfulness \cite{wang2022extracting}  & - & 82.7 & - \\
        \cmidrule{2-5}
        & Enhanced-Baseline  & 65.4 & 82.0 & 51.9 \\
        & METRE(Ours)  & 68.4 & 82.5 & 52.4 \\
        \bottomrule[2pt]
    \end{tabular}
    
    \caption{F1 score(\%) on the test set of TB-Dense, MATRES and UDS-T. For Enhanced-Baseline and METRE, we report the average results of 3 runs. *: Models are trained with additional resources. \dag: We re-run the code and report the result with the same evaluation metrics for a fair comparison.}
    \label{tab:tre_compare}
\end{table*}

Here, we first compare our model with state-of-the-art models and our baseline in Sec.\ref{sec:main result} and conduct an ablation study on Sec.\ref{sec:ablation study}. Then we analyze our performance on well-defined relations in Sec.\ref{sec:acc of nv} and discuss the effectiveness of Confusion Relation in Sec.\ref{sec:minority}. The efficient utilization of \textit{Vague} and the prediction interpretability are summarized in Sec.\ref{sec:fewshot} and Sec.\ref{sec:interpret}, respectively.

\subsection{Main Results}
\label{sec:main result}
Table~\ref{tab:tre_compare} compares the performance of our approach with previous works and our enhanced baseline. Our model makes a significant improvement based on the baseline model on all three benchmarks, and outperforms current state-of-the-art models by 0.6\% on TBDense and delivers a comparable result on MATRES. By modeling the latent composition of \textit{Vague}, our model can effectively leverage the hidden information about well-defined relations, leading to an enhanced understanding of temporal relations.


As \textit{Vague} accounts for nearly half of the training set of TB-Dense, which is far more than the proportions in MATRES and UDS-T, our model can explore more information from \textit{Vague} when trained on TB-Dense. Therefore, compared to baseline model, our model shows the largest improvement by 2.6\% F1 score on TB-Dense with BERT-Base as encoder, while 1.5\% and 1.0\% F1 improvements on MATRES and UDS-T, respectively. Experiments with RoBERTa-Large also show a similar trend, which indicates the efficacy of our model in extracting concealed information from \textit{Vague} to facilitate temporal relation prediction.


Our approach is still effective when provided with better event pair representations. As the encoder is changed from BERT-Base to RoBERTa-Large, our model shows a stable improvement on all benchmarks compared to the baseline. Due to our approach's exclusive focus on the classifier module, independent from the majority of previous works that concentrate on enhancing the encoder to acquire superior event pair representations, we believe that our approach can be effortlessly integrated into their models as a more effective classifier, and achieve better performance.

\subsection{Ablation Study}
\label{sec:ablation study}
\begin{table}[t]
    \small
    \center
    \setlength\tabcolsep{3.5pt}
    \begin{tabular}{l|ccc}
    \toprule
    & \makecell[c]{METRE\\w/o $CS$} & \makecell[c]{METRE\\w. Pnt} & METRE \\
    \midrule
    \textit{Before}      & 68.1 & 73.2 & 74.9 \\
    \textit{After}       & 68.6 & 72.6 & 73.1 \\
    \textit{Include}     & 37.6 & 36.3 & 38.5 \\
    \textit{Is included} & 36.9 & 35.5 & 39.2 \\
    \textit{Simultaneous}      & 0.0  & 7.8  & 5.1  \\
    \textit{Vague}       & 44.6 & 60.2 & 65.7 \\
    \midrule
    Micro-F1    & 64.1 & 66.0 & 67.9 \\
    \bottomrule
    \end{tabular}
    \caption{F1 scores of each relation and Micro-F1. }
    \label{tab:neg analyse}
    \vspace{-2.0mm}
\end{table}

The key of our approach is how we incorporate the mystery of \textit{Vague} into the design of $CS$.
To explore the impact of $CS$, we implement the model variant METRE w/o $CS$, where $CS$ is set to empty. Besides, we add the penalty on the relation $r\notin CS$ in another model variant METRE w. Pnt, to evaluate the negative effect of the penalty mechanism.
Experimental results are shown in Table~\ref{tab:neg analyse}.

\paragraph{$\mathbf{\textit{CS}}$ explores latent information} As mentioned in Sec.\ref{sec:pos_1}, $CS$ is a set that represents the inferred composition of \textit{Vague} based on the given text. Giving rewards to all relations in $CS$ is built on the assumption that the context contains the relevant information about these relations. To evaluate the effectiveness of $CS$, we conduct an experiment on the model variant METRE w/o $CS$ on TB-Dense. Compared to METRE w/o $CS$, METRE delivers much better results in every relation, which demonstrates $CS$ can effectively assist the model in better understanding every relation. For example, if the temporal relation of an event pair is designated as \textit{Vague} due to the ambiguity between \textit{Before} and \textit{Include}, our model can learn something about both two relations simultaneously. Therefore, after encouraging potential temporal relations to have higher possibilities, our model can efficiently use the latent information hidden under \textit{Vague} and hence achieve better performance.

\paragraph{Incorrect Penalty is Reduced} In our approach, we do not penalize any relation $r\notin CS$ since they may be the potential temporal relations between the event pair. To demonstrate the negative effect of penalty from \textit{Vague}, we add an extra loss $L_4$:
\begin{equation}
    \small
    L_4 = -log(\frac{exp(T)}{\sum_{r_i\in CS-r}exp(P(r_i))+exp(T)})
\end{equation}

Experiment results show that, the extra penalty of \textit{Vague} leads to the drop of F1 score on all relations except \textit{Simultaneous} (METRE w. Pnt does 0.33 more case of \textit{Simultaneous} correctly than METRE on the average of 3 runs). We attribute this negative effect to the incorrect penalty on some potential temporal relations between the event pair. For example, when both relation \textit{Before} and \textit{Simultaneous} could possibly be the temporal relation between an event pair, and \textit{Simultaneous} fails to be predicted into $CS$, its incorrect penalty may cause the model to misunderstand the meaning of \textit{Simultaneous}. This situation may be more serious in single-label classification-based methods, since they inevitably penalize every well-defined relation when faced with \textit{Vague}. However, our approach can easily avoid such situation, and reduce the incorrect penalty on potential temporal relations.

\subsection{Performance on Well-defined Relations}
\label{sec:acc of nv}
\begin{table}[t]
    \small
    \center
    \setlength\tabcolsep{2.3pt}
    \begin{tabular}{l|ccc|ccc}
    \toprule
    \multirow{2}{*}{\textbf{Relation}} & \multicolumn{3}{c|}{\textbf{F1}} & \multicolumn{3}{c}{\textbf{Recall}} \\ 
    & Base & METRE & Impr. & Base & METRE & Impr. \\
    \midrule
    \textit{Before}      & 72.4 & 74.9 & +2.5 & 70.2 & 79.3 & +9.1 \\
    \textit{After}       & 68.1 & 73.1 & +5.0 & 60.7 & 72.1 & +11.4 \\
    \textit{Include}     & 32.7 & 38.5 & +5.8 & 26.2 & 42.3 & +16.1 \\
    \textit{Is included} & 38.6 & 39.2 & +0.6 & 32.1 & 43.4 & +12.3 \\
    \textit{Simultaneous}       & 0.0  & 5.1  & +5.1 & 0.0  & 3.0  & +3.0 \\
    \textit{Vague}       & 68.8 & 65.7 & -3.1 & 76.7 & 63.6 & -13.1 \\
    \midrule
    Micro-F1    & 65.3 & 67.9 & +2.6 & 59.3 & 69.7 & +10.4 \\
    \bottomrule
    \end{tabular}
    \caption{F1 and Recall of each relation on TB-Dense.}
    \label{tab:rel recall}
    \vspace{-2.0mm}
\end{table}

$CS$ is composed of two parts: Top2 relations and confusion relation of Top1 relation. So when faced with \textit{Vague}, our model learns sufficient knowledge about at most three well-defined temporal relations. By leveraging such helpful information from \textit{Vague}, our model can have a better understanding of these relations. Therefore, we compare the performance of our model on every relation to baseline model, and report their F1 scores on TB-Dense in Table~\ref{tab:rel recall}. On all well-defined relations, our model shows higher F1 scores than baseline. This result proves that, with a stronger comprehension of temporal relations, our model is more confident in making decisions on a certain well-defined relation, and can discriminate them from each other better. 

The F1 score of \textit{Vague} is slightly decreased, and we found that it mainly results from its decrease in recall score, which is also reported in Table~\ref{tab:rel recall}. We can observe that, except \textit{Vague}, the recall scores increase greatly in all other relations, which is an encouraging indication that our model tends to recognize well-defined temporal relations between event pairs rather than predicting them as \textit{Vague}. Such a capability can enhance the utility of our model in downstream tasks, such as timeline construction, by providing more definitive decisions.

\subsection{Improvement in Minority Class}
\label{sec:minority}

\begin{figure}[t]
    \centering
    \hspace{-2mm}
    \subfigure[Dataset Information]{
        \label{fig:subfig:data info}
        \includegraphics[width=3.6cm, height=2.5cm]{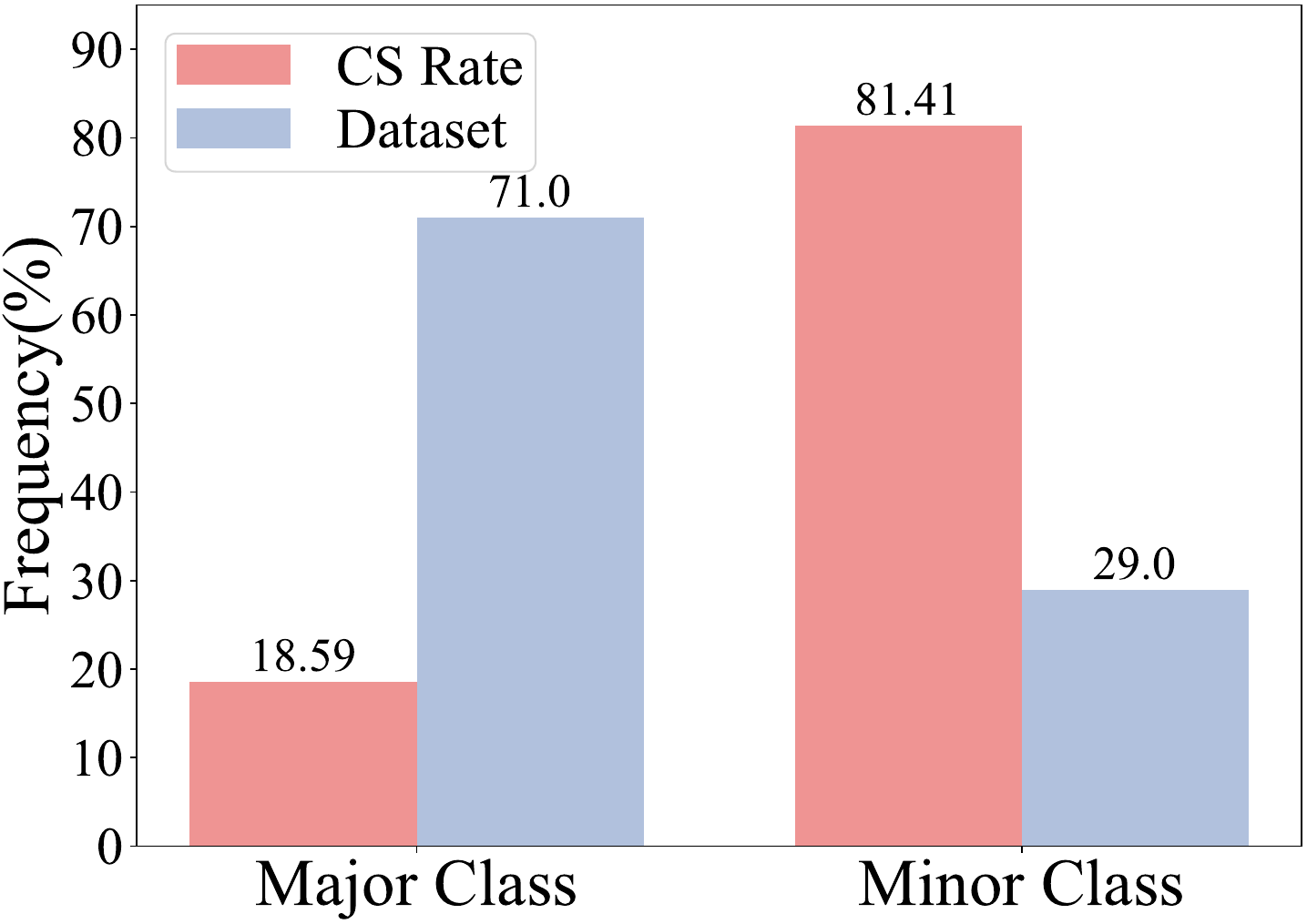}
        }
    \hspace{-2mm}
    \subfigure[Model Performance]{
        \label{fig:subfig:f1 score}
        \includegraphics[width=3.6cm, height=2.5cm]{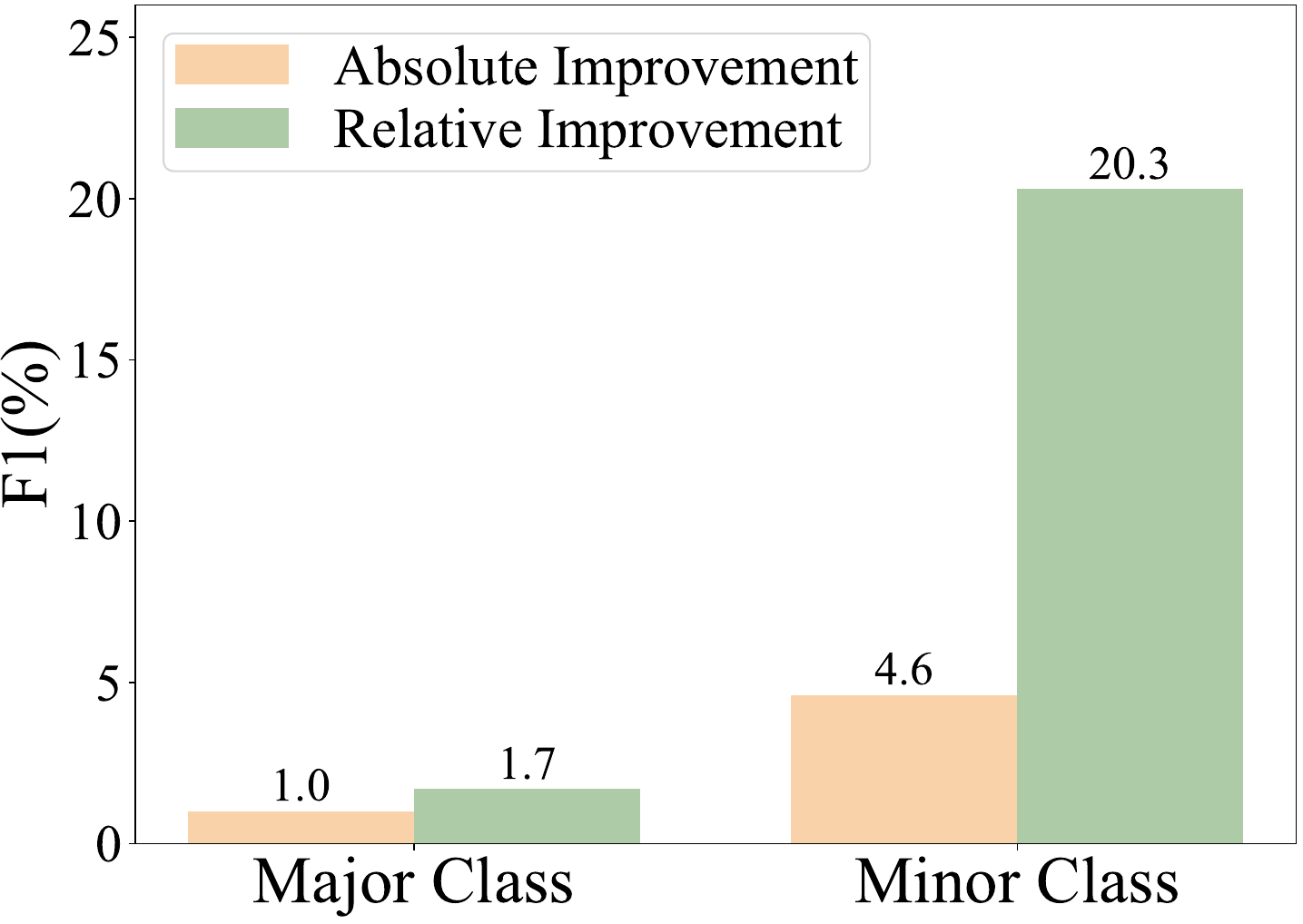}
       }

    \caption{Comparison between majority class and minority class. "CS Rate" indicates the class's proportion in $CS$ as a confusion relation, and "Dataset" indicates their proportion in the dataset.}
    \label{fig:minor analyse}
    \vspace{-2.0mm}
\end{figure}

When defining $CS$, we assume confusion relation is an important factor in causing the ambiguity of \textit{Vague}. Thus we carefully check what relations are brought into $CS$ as a confusion relation. We calculate the proportion of each class as a confusion relation in $CS$, and compare it to their proportion in the training set, as shown in Figure~\ref{fig:minor analyse}. The minority class, i.e. \textit{Include}, \textit{Is included} and \textit{Simultaneous}, which only account for 29.0\% in the training set, makes up of 81.4\% in the confusion relation part of $CS$. While the majority class, i.e. \textit{Before} and \textit{After}, only has a proportion of 18.6\%. This phenomenon may result from imbalanced label distribution in the dataset. As this exists in most temporal relation benchmarks, the model tends to predict the majority class with higher probability, so more minority class can be brought in by the majority class through confusion relations. Although it provides a valuable opportunity for the information of minority classes to be learned, we still need to figure out whether the confusion relations play an active role in explaining the intrinsic composition of \textit{Vague}. 

Therefore, we compare the performance of both majority class and minority class to our baseline model. Table~\ref{fig:subfig:f1 score} shows that, the minority class achieves a greater improvement of 4.6\% F1 score and 20.3\% relative improvement compared to baseline, while majority class only achieves 1.0\% and 1.7\% improvement respectively. The positive effect on the performance of minority class demonstrates the effectiveness of confusion relations, which can reasonably explain where the uncertainty of an event pair's temporal relationship comes from. Besides, by providing more opportunities for our model to explore and learn the knowledge of minority classes, it enables us to better understand the minority class, and alleviates the imbalanced label distribution issue to some extent.

\subsection{Utilizing \textit{Vague} Efficiently}
\label{sec:fewshot}
\begin{figure}
    \centering
    \includegraphics[width=5.0cm, height=3.8cm]{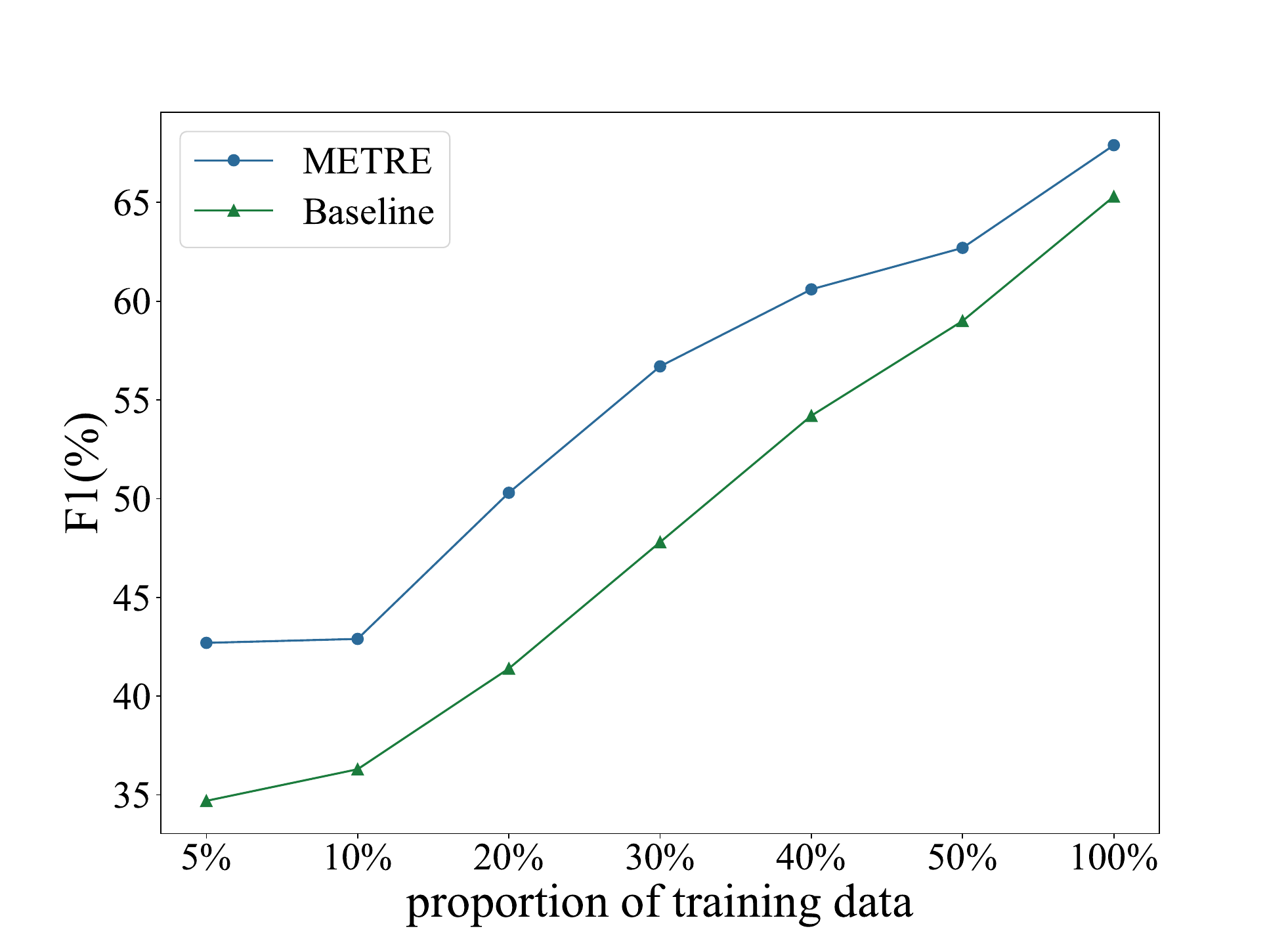}
    \caption{Results of few shot learning on TB-Dense.}
    \label{fig:fewshot}
\end{figure}

Event pairs with temporal relation \textit{Vague} contain sufficient information about well-defined relations. In our approach, our model learns the meaning of a well-defined relation not only from itself, but also from \textit{Vague}. The importance of supplementary information provided by \textit{Vague} could be more obvious when the training resources are limited. Therefore, to evaluate our model's capability in effectively exploring sufficient information from training data, we conduct experiments with variable amounts of training data, ranging from 5\% to 50\%, while keeping the test set unchanged. In low resources scenarios, our model outperforms baseline model by 7.1\% on average F1 score. Figure~\ref{fig:fewshot} illustrates the experiment results of the F1 scores of baseline model and our model on TB-Dense. 

We can observe that, our model stably outperforms baseline model in every portion of the training set. This result indicates that our model is more capable of capturing the information in \textit{Vague} efficiently and making full use of it. When the smaller size of the training set is used, such as 5\% to 30\%, our model shows a larger difference from baseline. This phenomenon mainly results from the different amounts of information models can obtain. Since there are only a few well-defined relations in the small training set, single-label classification-based models are struggling to recognize them correctly. However, our model's capability in leveraging \textit{Vague} to learn knowledge of well-defined relations plays an important role, which helps to capture effective information from the limited training set more comprehensively and efficiently.

In addition, due to the imbalanced label distribution, minority classes like \textit{Include} and \textit{Is included} are hardly included in the small training set. Therefore, single-label classification-based methods may have difficulty in making predictions on these relations. Our method, on the other hand, can explore their information through a large amount of \textit{Vague}, and achieves higher average F1 scores of 4.7\% and 6.2\% on \textit{Include} and \textit{Is included} respectively. 

\subsection{Interpretability of \textit{Vague}}
\label{sec:interpret}

\begin{table}[t]
    \small
    \center
    \setlength\tabcolsep{4.2pt}
    \begin{tabular}{l|ccc|ccc}
    \toprule
    \multirow{2}{*}{\textbf{Precision}} & \multicolumn{3}{c|}{\textbf{Absolute Value}} & \multicolumn{3}{c}{\textbf{Relative Value}} \\
             & Top1 & Top2 & Top3 & Top1 & Top2 & Top3  \\
    \midrule
    Random   & 60.0 & 30.0 & 10.0 & 0.0  & 0.0   & 0.0   \\
    Baseline & 76.9 & 48.7 & 19.9 & 28.2 & 62.3  & 99.0  \\
    METRE     & 82.1 & 63.1 & 28.2 & 36.8 & 110.3 & 182.0 \\
    \rowcolor{Gray} Improve. & +5.2 & +14.4 & + 8.3 & *1.30 & *1.77 & *1.84 \\
    \bottomrule
    \end{tabular}
    \caption{Absolute precision and relative precision on Top$K$ prediction of \textit{Vague}.}
    \label{tab:vague precision}
\end{table}

Treated as a multi-label classification task, our model is able to predict relations composition of \textit{Vague}. To better evaluate the prediction accuracy, we check the consistency between our prediction of \textit{Vague} and the annotation results from humans. As we mentioned in Section~\ref{sec:udst}, every event pair's relation is determined by 3 annotators, so we can calculate the overlap of the prediction of our model with the 3 different annotation results. Specifically, if we make a correct prediction on \textit{Vague}, we choose the Top$K$ relations and judge whether these $K$ relations all conform to the annotation results. The precision scores are shown in Table~\ref{tab:vague precision}. 

Our model outperforms baseline by 5.2\%, 14.4\% and 8.3\% in Top1/2/3 precision respectively, demonstrating a greater capability of our model in predicting the intrinsic composition of \textit{Vague}. Given the precision of random ranking, we calculate the relative precision value as $Pr_{ours}/Pr_{random}-1$ and $Pr_{base}/Pr_{random}-1$. As $K$ increases, our model shows a larger improvement in relative precision than baseline, from 1.30 times to 1.84 times, which indicates that our advantage is more obvious when the model is asked to provide full information of \textit{Vague}.

\section{Conclusion}
In this work, we investigate the underlying meaning of the special relation \textit{Vague} in the temporal relation extraction task. A novel approach is proposed to effectively utilize the latent connection between \textit{Vague} and well-defined relations, helping our model understand the meaning of each relation more accurately. Experiments show that our model achieves better or comparable performance compared to previous state-of-the-art methods
on three benchmarks, indicating the effectiveness of our approach. Extensive analyses further demonstrate our model's advantage in using training data more comprehensively and making our predictions more interpretable. 

\section*{Limitations}
There are only 3 well-defined relations in MATRES and none of them have confusion relation, as these relations are only determined by the start-point relations. Thus the diversity of $POS_{Vague}$ is constrained due to the absence of confusion relation and the small size of well-defined relations, which further influences its effectiveness in assisting the model's understanding of \textit{Vague}. Besides, there is not a proper confusion relation for \textit{Simultaneous} currently. Therefore, we still need to figure out a better constitution of $POS_{Vague}$ and further improve the performance of our model. 


\bibliography{custom}
\bibliographystyle{acl_natbib}

\appendix
\section{Dataset Statistics}
\label{app:data statistics}
\begin{table}[h]
    \small
    \center
    \begin{tabular}{l|ccc}
    \toprule
    Dataset & TB-Dense & MATRES & UDS-T \\ 
    \midrule
    \textit{Before} & 808 & 5483 & 2427 \\
    \textit{After} & 674 & 3651& 323 \\
    \textit{Include} & 206 & - & 523 \\
    \textit{Is included} & 273 & - & 514 \\
    \textit{Simultaneous} & 59 & 376 & 86 \\
    \textit{Vague} & 2012 & 1320 & 1217 \\
    \midrule
    \textit{Vague} Proportion & 49.9\% & 12.2\% & 23.9\% \\
    \bottomrule
    \end{tabular}
    \caption{Data statistics for the training set of TB-Dense, MATRES and UDS-T}
    \label{tab:training dataset info}
\end{table}

\begin{table}[h]
    \small
    \center
    \begin{tabular}{l|ccc}
    \toprule
    Dataset & TB-Dense & MATRES & UDS-T \\ 
    \midrule
    \textit{Before} & 1348 & 6853 & 3316 \\
    \textit{After} & 1120 & 4548& 439 \\
    \textit{Include} & 276 & - & 784 \\
    \textit{Is included} & 347 & - & 693 \\
    \textit{Simultaneous} & 93 & 470 & 119 \\
    \textit{Vague} & 2012 & 1631 & 1682 \\
    \midrule
    \textit{Vague} Proportion & 47.7\% & 12.1\% & 23.9\% \\
    \bottomrule
    \end{tabular}
    \caption{Data statistics for the full set of TB-Dense, MATRES and UDS-T}
    \label{tab:full dataset info}
\end{table}

\section{UDS-T Mapping Rules}
\label{app:udst map}

\begin{table}[h]
    \center
    \begin{tabular}{cc|c}
    \toprule
     start-point & end-point & relation \\ 
    \midrule
     before & before & \textit{Before} \\
     before & equal  & \textit{Include} \\
     before & after  & \textit{Include} \\
     equal  & before & \textit{Is included} \\
     equal  & equal  & \textit{Simultaneous} \\
     equal  & after  & \textit{Include} \\
     after  & before & \textit{Is included} \\
     after  & equal  & \textit{Is included} \\
     after  & after  & \textit{After} \\
    \bottomrule
    \end{tabular}
    \caption{udst mapping rules}
    \label{tab:udst map}
\end{table}

In UDS-T, instead of determining the temporal relation between an event pair explicitly, annotators are required to determine the relative timeline between every event pair. There are only three temporal relations between a time-point pair: \textit{before}, \textit{after} and \textit{equal}, and we can deduce the temporal relation according to the relation definition in TB-Dense, as shown in Table~\ref{tab:udst map}.

Every event pair in the training set of UDS-T is annotated by only one person, where inconsistency and ambiguity will not happen and \textit{Vague} is not considered in the training set. The validation set and test set, on the other hand, are annotated by three different people per event pair. Therefore, we use majority voting to decide which temporal relation of an event pair should be. Specifically, if two or three people make the same decision, we hold it as the temporal relation between this event pair, while if three annotators make totally different decisions, we view it as \textit{Vague}, and take all these three decisions as the possible relations between this event pair. Since only the validation set and test set of UDS-T contain \textit{Vague}, we split them into the new training/validation/test set to fit the temporal relation task in our work, with the proportion of 80\%/10\%/10\% respectively.  

\end{document}